\documentclass[letterpaper, 10 pt, conference]{ieeeconf}  % Comment this line out if you need a4paper

\IEEEoverridecommandlockouts                              % This command is only needed if
\usepackage{graphicx} % for pdf, bitmapped graphics files
\usepackage{amsmath} % assumes amsmath package installed
\usepackage{xcolor}
\usepackage{booktabs}
\usepackage{adjustbox}

\makeatletter
\let\MYcaption\@makecaption
\makeatother
\usepackage[font=footnotesize,labelformat=simple]{subcaption}
\makeatletter
\let\@makecaption\MYcaption
\makeatother

\makeatletter
\let\NAT@parse\undefined
\makeatother
\usepackage[hidelinks]{hyperref} % should be loaded last

\usepackage[noabbrev,capitalise,sort&compress,nameinlink]{cleveref} % should be loaded even after hyperref

\newcommand{\network}{\mbox{PWOC-3D}} % Just change the name here if we decide on a new network name.
\newcommand{\vectr}[1]{\mathrm{\textbf{#1}}}
\newcommand{\ilt}{$\vectr{I}_L^t$} % convenience commands for image symbols
\newcommand{\irt}{$\vectr{I}_R^t$}
\newcommand{\ilT}{$\vectr{I}_L^{t+1}$}
\newcommand{\irT}{$\vectr{I}_R^{t+1}$}
\newcommand{\clt}{$_l\vectr{c}_L^t$}
\newcommand{\crT}{$_l\vectr{c}_R^{t+1}$}
\newcommand{\crt}{$_l\vectr{c}_R^{t}$}
\newcommand{\clT}{$_l\vectr{c}_L^{t+1}$}

\title{\LARGE \bf
PWOC-3D: Deep Occlusion-Aware End-to-End Scene Flow Estimation
}

\author{Rohan Saxena$^{1,2}$ \quad Ren\'e Schuster$^{1}$ \quad Oliver Wasenm\"uller$^{1}$ \quad Didier Stricker$^{1,3}$% <-this % stops a space
\thanks{$^{1}$German Research Center for Artificial Intelligence - DFKI,
        Kaiserslautern, Germany,
        {\tt\small firstname.lastname@dfki.de}}
\thanks{$^{2}$Birla Institute of Technology and Science - BITS Pilani, India}
\thanks{$^{3}$University of Kaiserslautern - TUK, Kaiserslautern, Germany}%
}

\begin{document}
\bstctlcite{IEEEexample:BSTcontrol}
\maketitle
\thispagestyle{empty}
\pagestyle{empty}

%%%%%%%%%%%%%%%%%%%%%%%%%%%%%%%%%%%%%%%%%%%%%%%%%%%%%%%%%%%%%%%%%%%%%%%%%%%%%%%%
\begin{abstract}
    In the last few years, convolutional neural networks (CNNs) have demonstrated increasing success at learning many computer vision tasks including dense estimation problems such as optical flow and stereo matching. However, the \textit{joint} prediction of these tasks, called scene flow, has traditionally been tackled using slow classical methods based on primitive assumptions which fail to generalize. The work presented in this paper overcomes these drawbacks efficiently (in terms of speed and accuracy) by proposing \network, a compact CNN architecture to predict scene flow from stereo image sequences in an end-to-end supervised setting.
Further, large motion and occlusions are well-known problems in scene flow estimation. \network{} employs specialized design decisions to explicitly model these challenges. In this regard, we propose a novel self-supervised strategy to predict occlusions from images (learned without any labeled occlusion data). Leveraging several such constructs, our network achieves competitive results on the KITTI benchmark and the challenging FlyingThings3D dataset. Especially on KITTI, \network{} achieves the second place among end-to-end deep learning methods with 48 times fewer parameters than the top-performing method.
\end{abstract}

%%%%%%%%%%%%%%%%%%%%%%%%%%%%%%%%%%%%%%%%%%%%%%%%%%%%%%%%%%%%%%%%%%%%%%%%%%%%%%%%

\section{INTRODUCTION}
In robot navigation, particularly in an autonomous driving pipeline, estimating the motion of other traffic participants is one of the most crucial components of perception. Scene flow is one such reconstruction of the complete 3D motion of objects in the world. Due to the rich perceptual information it provides, it serves as a foundation for several high-level driving tasks in Advanced Driver Assitance Systems (ADAS) and autonomous systems.

Owing to the increased complexity of a full 3D reconstruction, the projection of this 3D motion on the image plane (called optical flow) often serves as an approximate proxy for motion sensing and perception in intelligent vehicles. Consequently, individual components of scene flow, namely optical flow and stereo disparity, have received significantly more attention in computer vision research.

However, the joint estimation of these tasks as scene flow has multiple benefits: Firstly, stereo matching is essentially a special case of optical flow where pixel matching is constrained along the epipolar line. Using a shared representation for these related tasks can result in fewer trainable parameters with a smaller memory footprint and faster training/inference with fewer computational resources. Secondly, the principle of multi-task learning \cite{multitasklearning,kendall2018multi} states that jointly training this shared representation improves generalization by leveraging domain-specific information across tasks, with the training signals for one task serving as an inductive bias for the other.

Our scene flow approach also offers the advantages of robustness and speed over earlier approaches, both of which are critical for deployment in intelligent vehicles. Firstly, classical methods (particularly variational techniques) are based on data consistency assumptions like constancy of brightness or gradient across images, or the smoothness and constancy of motion within a small region. Such primitive assumptions fail to generalize due to varying illumination (shadows or lighting changes), occlusions or large displacements; all of which are prevalent on the dynamic road scenes. Since such scenes are a common scenario for intelligent vehicles, these methods are rendered ineffective on them. In contrast, our network does not make \textit{any} such assumptions and can be used to estimate scene flow in a wide variety of environments with complex motions. Secondly, autonomous driving systems and embedded devices require real-time estimation of scene flow. The traditional approaches are iterative and can take anywhere between 1-50 minutes to process a set of images. In comparison, our end-to-end CNN can perform the same task with a single forward pass of the network in less than 0.2 seconds.

\begin{figure}
	\centering
	\begin{subfigure}[t]{0.49\linewidth}
		\includegraphics[width=\linewidth]{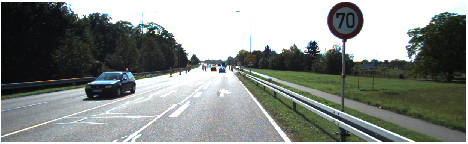}
		\caption{Reference input image: \ilt.}
		\label{fig:teaser-ref}
	\end{subfigure}
    \begin{subfigure}[t]{0.49\linewidth}
		\includegraphics[width=\linewidth]{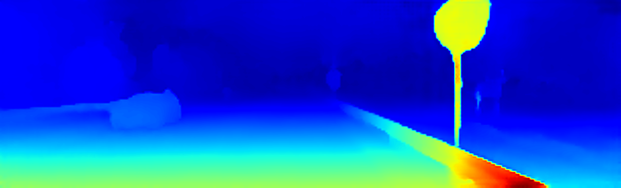}
		\caption{First disparity prediction $d_{0\Theta}$.}
		\label{fig:disp}
	\end{subfigure}
    \begin{subfigure}[t]{0.49\linewidth}
		\includegraphics[width=\linewidth]{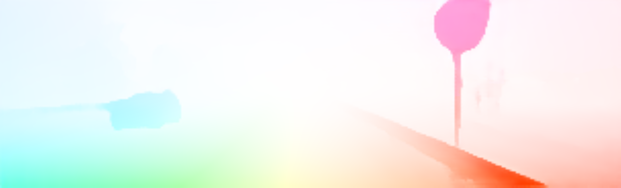}
		\caption{Optical flow prediction: $\vectr{u}_\Theta$.}
		\label{fig:teaser-flow}
	\end{subfigure}
    \begin{subfigure}[t]{0.49\linewidth}
		\includegraphics[width=\linewidth]{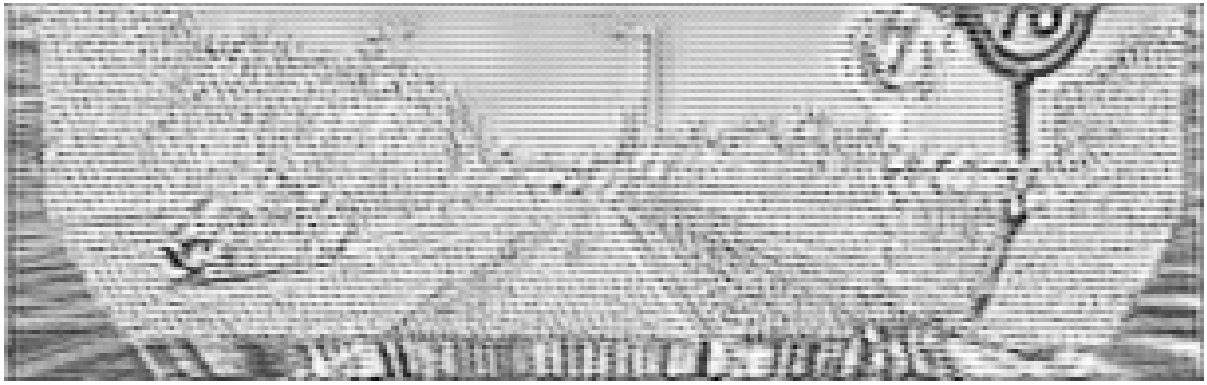}
		\caption{Soft occlusion mask: $\vectr{o}_L^{t+1}$.}
		\label{fig:maskflow}
	\end{subfigure}
%	\begin{subfigure}[c]{0.49\linewidth}
%		\includegraphics[width=\linewidth]{media/teaser/new/od0.png}
%		\caption{Soft occlusion mask: $\vectr{o}_R^{t}$.}
%		\label{fig:maskd0}
%	\end{subfigure}
%	\begin{subfigure}[c]{0.49\linewidth}
%		\includegraphics[width=\linewidth]{media/teaser/new/od1.png}
%		\caption{Soft occlusion mask: $\vectr{o}_R^{t+1}$.}
%		\label{fig:maskd1}
%	\end{subfigure}
	\caption{Example predictions of our \network{} network on the KITTI benchmark. \network{} uses CNNs to predict scene flow in an efficient end-to-end fashion. The soft occlusion map \subref{fig:maskflow} is  predicted by our novel self-supervised occlusion reasoning mechanism, which is leveraged to improve scene flow estimates.}
	\label{fig:teaser}
\end{figure}

Our network also contains specialised constructs to handle specific challenges in a typical scene flow pipeline. Firstly, inspired by the work of PWC-Net \cite{sun2018pwc} for optical flow, we employ a coarse-to-fine estimation scheme using a spatial pyramid and warp image features at each pyramid level using the intermediate flow estimate from the previous level to handle large motion. Secondly, we propose a novel self-supervised strategy to predict dense occlusion maps from images without using any labelled occlusion data and use these to improve scene flow estimates (cf. \cref{fig:teaser}). To the best of our knowledge, ours is the first method to reason about occlusion using a single flow prediction and without any occlusion ground truth. Previous methods \cite{hur2017mirrorflow,meister2018unflow,wang2018occlusion} require at least bidirectional flow (forward and backward preditions) to model occlusion, thus our network reduces the effort by half compared to these methods.

Our network design demonstrates that embedding vision techniques, which leverage the underlying domain-knowledge of the problem and geometry of the scene, within differentiable CNNs produces better results than either approach has been able to single-handedly achieve. We demonstrate the performance of our method on the KITTI benchmark, where our method achieves the second highest accuracy among end-to-end CNN methods with 48 times fewer parameters than the top-performing method \cite{occlusionssceneflow}.

\section{RELATED WORK}
\textbf{Conventional Scene Flow.} 
First scene flow approaches were inspired by variational methods for optical flow \cite{hornschunck}.
These approaches frame scene flow estimation as global minimization problem and optimize it using variational calculus. 
The total energy or cost consists of a data term and a smoothness term, which is then optimized iteratively using the Euler-Lagrange equations. The data term typically models a means of 3D reconstruction of the scene by incorporating a cost based on the similarity of pixel intensities at the reference image and the other images warped towards the reference image using flow predictions. There are various works in this area with different types of underlying assumptions, such as \cite{huguet2007variational, herbst2013rgb, basha2013multi, park2012tensor, zhang2012dense, ferstl2014atgv, quiroga2014dense}.

To overcome the data consistency assumptions of variational methods, various techniques have explored segmenting the scene into rigid planes and estimating piecewise rigid scene flow. The work of \cite{menze2015object} decomposes the scene into such piecewise rigid planes (superpixels) and employs a discrete-continuous CRF to optimize scene flow. \cite{menze2018osf} further combines the scene flow model with discrete optical flow estimates \cite{menze2015discrete} to handle large motion. The work of \cite{neoral2017object} improves this model by propagating the object labels and their motion information from the previous frame to produce temporally consistent scene flow. The work of \cite{lv2016continuous} also uses superpixel segmentation and formulates scene flow using a factor graph, which enables decomposition of the problem into photometric, geometric and smoothing constraints.
Some methods have also used a similar formulation of the problem while exploiting multiple views. For instance, \cite{vogel20153d} also employs rigid plane segmentation, but in a multi-frame setting, and performs segment-level occlusion reasoning. The work of \cite{taniai2017fast} decomposes multi-frame scene flow into rigid and non-rigid optical flow and stereo matching, combined with ego-motion estimation and motion segmentation.
All methods in this category are greatly restricted in applicability due to the rigid motion formulation, which suits only the KITTI dataset well (since its ground truth was constructed by modelling dynamic objects as rigid). On the other hand, our method can be applied in a general scenario with non-rigid and non-planar objects moving in arbitrary 3D paths. This is particularly important for intelligent vehicles which encounter non-rigid objects such as motorbikes, cyclists and pedestrians in dynamic road scenes.
SceneFlowFields \cite{schuster2018sceneflowfields} uses a sparse-to-dense approach instead of rigid-planar assumptions for regularization to achieve a better generalization. Though the accuracy of this methods is competitive, the speed is still far from real time.

\textbf{Semantic Scene Flow.}
More recent methods explore CNNs for semantic segmentation over superpixel segmentation. The work of \cite{behl2017bounding} studied different granularities of instance recognition (including segmentation) using CNNs and explored how they could be employed to improve scene flow predictions from a CRF model. \cite{ren2017cascaded} similarly leverages instance-level semantic segmentation cues from a CNN to improve piecewise-rigid scene flow estimates using a cascade of CRFs. This class of methods will be heavily biased on the instance-level recognition dataset that is used to train the CNNs. In the settings described, the instances were obtained from the same dataset as the scene flow benchmark (KITTI), which is not the case in general.

\textbf{End-to-end CNN for Optical Flow.} The incorporation of a spatial pyramid and warping at different pyramid levels in an end-to-end CNN for optical flow estimation was first introduced in SPyNet \cite{spynet}. PWC-Net \cite{sun2018pwc} built upon SPyNet's pipeline by replacing the latter’s image pyramid with a feature pyramid learned using a CNN. A cost volume was used to predict optical flow instead of plain features, and an additional network was used to refine predictions from the last pyramid level. Our \network{} uses the PWC-Net architecture as a skeleton, but differs from it in several ways. Firstly, \network{} reasons about the full 3D motion of objects rather than just 2D optical flow. This is made possible through several key design decisions: We construct four image pyramids (one for each image in the stereo sequence), we define 1D (for disparity) and 2D (for optical flow) versions of the warping and cost volume operations in our network. The 1D operations leverage the epipolar constraint for rectified stereo images to limit computation. Secondly, we employ a feature pyramid network (FPN) \cite{lin2017feature} to construct the feature pyramids instead of PWC-Net's generic CNN feature extractor (as illustrated in \cref{fig:fpn}), with significant improvement in results. Thirdly, \network{} explicitly reasons about occlusion via our novel self-supervised method of predicting occlusion directly from images (without any occlusion ground truth), and exploits this understanding to improve scene flow predictions. The entire \network{} pipeline is described in detail in the \cref{sec:method}.

\textbf{End-to-End CNN for Scene Flow.} Currently, there are only a couple of other end-to-end CNN architectures published for scene flow. The first was proposed alongside the FlyingThings3D \cite{mayer2016large} dataset primarily as a proof-of-concept of the utility of the dataset. This network contained roughly three times the number of trainable parameters of FlowNet \cite{flownet}. In contrast, our method outperforms it while being smaller than a single FlowNet model. The second end-to-end CNN work was presented in \cite{occlusionssceneflow}. This network used three separate processing pipelines to predict optical flow, initial and final disparities respectively. It was able to demonstrate competitive performance on the KITTI benchmark, although with the parameters of ten FlowNet models. In contrast, our fast and compact network is a close second place to this method with 48 times fewer parameters.

\begin{figure}
	\centering
	\begin{subfigure}[b]{0.49\linewidth}
		\includegraphics[width=\linewidth]{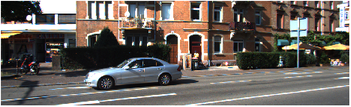}
		\caption{Reference input image: \ilt.}
		\label{fig:warp-ref}
	\end{subfigure}
    \begin{subfigure}[b]{0.49\linewidth}
		\includegraphics[width=\linewidth]{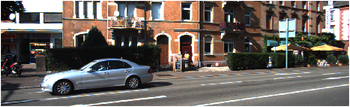}
		\caption{Left image from $t+1$: \ilT.}
		\label{fig:inp}
	\end{subfigure}
    \begin{subfigure}[b]{0.49\linewidth}
		\includegraphics[width=\linewidth]{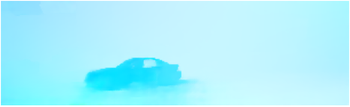}
		\caption{Our predicted optical flow from \ilt{} to \ilT.}
		\label{fig:warp-flow}
	\end{subfigure}
    \begin{subfigure}[b]{0.49\linewidth}
		\includegraphics[width=\linewidth]{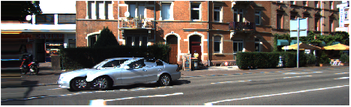}
		\caption{\ilT warped towards \ilt{} using our predicted optical flow.}
		\label{fig:out_pred}
	\end{subfigure}
	\caption{The adverse effect of occlusion on the warping operation. In \subref{fig:out_pred}, there are 2 cars visible: The car on the right is the `true' car. The part of the road which is visible in \ilt{} and occluded by the car in \irt{} is static, due to which the occluding area of the car is reproduced incorrectly from \ilt.}
	\label{fig:warp}
\end{figure}

\textbf{Occlusion Handling for Flow.}
MirrorFlow \cite{hur2017mirrorflow} predicted bidirectional flow using variational methods and used it to warp both images towards each other. A forward-backwards consistency check was imposed on these warps. Areas which did not pass this check were considered occluded. This led them to predict consistent occlusion maps in both directions. UnFlow \cite{meister2018unflow} used a very similar formulation of the problem: Bidirectional flow estimation, forward-backward consistency check, occlusion estimation. The difference here was that a FlowNet was used to predict flow instead of variational methods, and occluded areas were masked from contributing to the reconstruction loss function. Wang et al. \cite{wang2018occlusion} also used the same basic pipeline as UnFlow. This work proposed a different method of predicting occlusion maps based on warping a constant grid using the predicted flow.

All previous methods predicted occlusion using bidirectional flow. In contrast, \network{} estimates occlusions in three images \irt, \ilT, \irT{} without any labeled occlusion data while computing only the forward direction flow.

\section{METHOD} \label{sec:method}
\begin{figure}[t]
	\centering
	\includegraphics[width=1\linewidth]{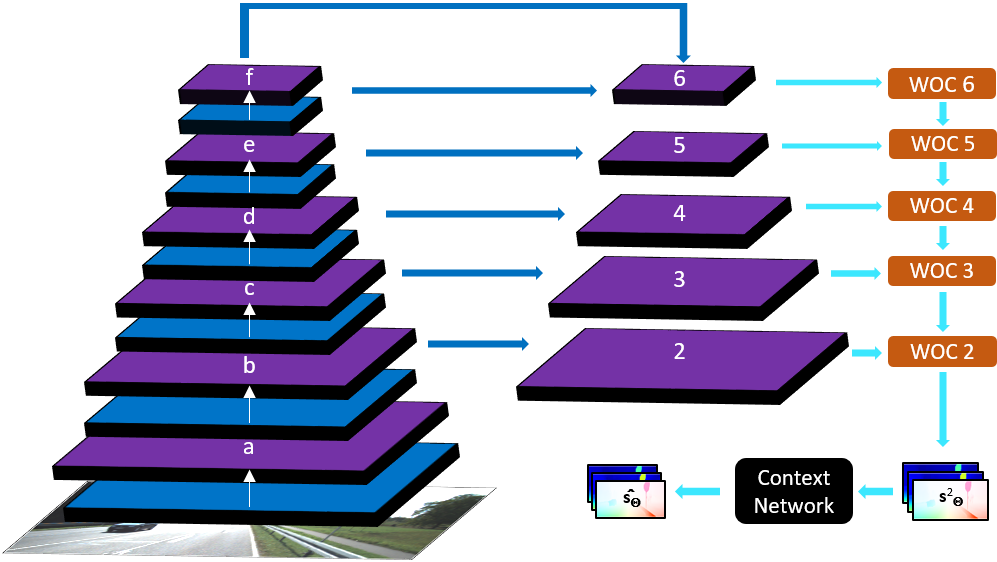}
	\caption{Visualization of the flow of information across pyramid levels in our entire \network{} pipeline. PWC-Net uses only the levels $b, c, d, e, f$ as a feature pyramid, while we use levels $2, 3, 4, 5, 6$. The orange boxes represent warping, occlusion estimation, cost volume computation, and scene flow prediction for one level of the pyramid as shown in \cref{fig:pipeline}.}
	\label{fig:fpn}
\end{figure}

\begin{figure*}[t]
    \centering
    \includegraphics[width=\linewidth]{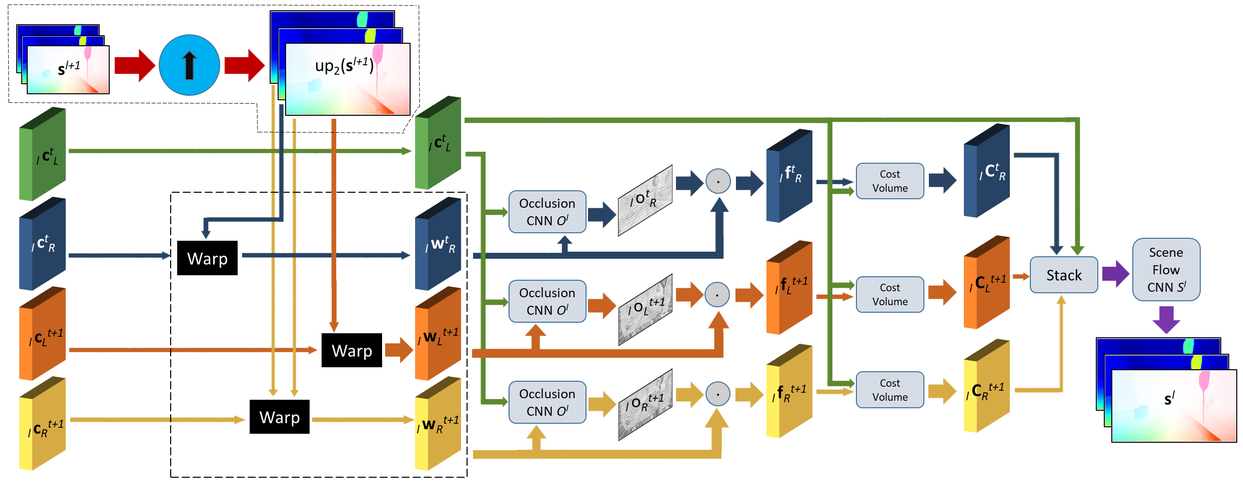}
    \caption{An overview of the inference pipeline of \network{} at pyramid level $l$. The operations within the dashed boundary denote the warping operations, which are present at every pyramid level except the topmost.}
    \label{fig:pipeline}
\end{figure*}

Our PWOC-3D pipeline involves extracting a feature pyramid for each of the four images \ilt, \irt, \ilT, \irT. The features of \irt, \ilT, \irT{} at a particular pyramid level (except the top, i.e. lowest resolution) are warped towards the features of \ilt{} using the flow estimates from the upper pyramid level. Based on the warped features, occlusion maps are predicted for \irt, \ilT, \irT. A cost volume is then constructed using the features of \ilt{} and each of the warped features of \irt, \ilT, \irT{} considering the predicted occlusions. Afterwards, a scene flow estimator network is used to predict scene flow using these cost volumes. Finally, a context network with dilated convolutions is used to refine the scene flow estimates. The complete overview of the end-to-end architecture is given in \cref{fig:fpn}. \Cref{fig:pipeline} shows a detailed view of the pipeline at a particular pyramid level $l$.

\textbf{Feature Pyramids.} In PWC-Net \cite{sun2018pwc}, a simple feedforward strided CNN is used to construct feature pyramids for both the input images. However, using the different feature maps of a generic CNN in this manner is not an optimal strategy. This is because the high-resolution feature maps from the first few layers of the network contain well-localized, but semantically weak features; while low-resolution maps from deeper layers contain processed and semantically strong features which are not well-localized with the original image due to strided subsampling of the convolution operation \cite{schuster2019sdc}.

The FPN \cite{lin2017feature} work proposes to overcome this problem by incorporating additional connections in the network as shown in \cref{fig:fpn}. In addition to the backbone bottom-up pathway (which computes a feature hierarchy as in a standard CNN), top-down pathways and lateral residual connections are introduced. The top-down pathway produces higher resolution semantically stronger feature maps, while the lateral skip connections (from layers lower in the pipeline) reinforce localization of features with respect to the input images. Combined, this mechanism leads to the feature map at each pyramid level being well-localized and semantically strong.

Since our architecture uses a coarse-to-fine estimation approach, especially with predictions from higher up in the pyramid being used in lower levels, consistency of semantic strength and spatial localization of the features across levels becomes particularly important. Thus, our work explores the role of FPN-like connections in the \network{} pipeline.

For each of the 4 input images \ilt, \irt, \ilT{}, \irT, we use the same network (from \cref{fig:fpn}) to construct 4 six-layered feature pyramids (denoted as $\vectr{c}_L^{t}, \vectr{c}_R^{t}, \vectr{c}_L^{t+1}, \vectr{c}_R^{t+1}$ respectively), with each subsequent pyramid level having half the resolution (in each dimension) of its predecessor, so that the subsampling factor at layer $l$ is $2^l$ for each dimension. We start processing from the topmost level and continue the coarse-to-fine estimation scheme until feature pyramid level 2. That is, \network{} produces scene flow estimations at $1/4$th of the input resolution in each dimension. We upsample the prediction using bilinear interpolation to obtain full-scale scene flow.

\textbf{Warping.}
At every pyramid level $l$, the feature maps of \irt, \ilT, \irT{} (denoted as \crt, \clT, \crT{} respectively) are warped towards the reference image \ilt. Let the scene flow estimate at level $l$ be denoted as $\vectr{s}^l = (u^l, v^l, d^l_0, d^l_1)^\top$, the images are warped as follows:
\begin{itemize}
    \item \crt{} is warped towards \ilt{} using the disparity $d^{l+1}_0$, a 1D warping:
    \begin{equation}
        {_l}\vectr{w}_{R}^t(\vectr{x}) = \text{}_l\vectr{c}_R^t\left(\left(x - \text{up}_2\left(d^{l+1}_0\right)(\vectr{x}), \text{ }y\right)^\top\right),
    \end{equation}
    where $\text{up}_2(d^{l+1}_0)$ denotes the predicted disparity map from level $l+1$ which has been upsampled by a factor of 2 using bilinear interpolation, and $\vectr{x} = (x, y)^\top$ denotes the pixel index.
    \item \clT{} is warped towards \ilt{} using optical flow $\vectr{u}^{l+1} = (u^{l+1}, v^{l+1})^\top$, a 2D warping:
    \begin{equation}
        {_l}\vectr{w}_{L}^{t+1}(\vectr{x}) = \text{}_l\vectr{c}_L^{t+1}\left(\vectr{x} + \text{up}_2\left(\vectr{u}^{l+1}\right)(\vectr{x})\right).
    \end{equation}
    \item \crT{} is warped towards \ilt{} using optical flow $\vectr{u}^{l+1} = (u^{l+1}, v^{l+1})^\top$ and disparity $d_1^{l+1}$, a modified 2D warping:
    \begin{equation}
        \begin{split}
            &{_l}\vectr{w}_{R}^{t+1}(\vectr{x}) = \\
            & \text{}_l\vectr{c}_R^{t+1}\left(\vphantom{\left(y + \text{up}_2\left(v^{l+1}\right)(\vectr{x})\right)^\top}\left(x - \text{up}_2\left(d_1^{l+1}\right)(\vectr{x}) + \text{up}_2\left(u^{l+1}\right)(\vectr{x}), \right.\right. \\
           & \left.\left. y + \text{up}_2\left(v^{l+1}\right)(\vectr{x})\right)^\top\right).
        \end{split}
    \end{equation}
\end{itemize}

\textbf{Occlusion Mechanism.}
Occlusion, which is omnipresent in real-world dynamic scenes, plays an important role in the estimation of scene flow. Firstly, it leads to incorrect matching costs being computed since the object of interest is occluded from view. Secondly, the lack of information about the occluded area can throw off a naïve method because tracking the occluded pixels directly is impossible, and we must leverage other information to estimate this occluded motion. Thirdly, a considerable area of the reference image is occluded from view when it moves out of the field of view of the camera due to motion of the camera itself. This ego-motion is an inherent characteristic of autonomous driving systems. Thus, failing to account for occlusion has significant drawbacks.
Occlusion also has an adverse effect on the (1D and 2D) warping operation, as illustrated in \cref{fig:warp}.

\network{} employs a novel strategy to handle occlusion by learning an occlusion model conditioned on the input images. The occlusion mechanism is explained using \irt, but also applies in an analogous manner to the images \ilT{} and \irT. Specifically, occlusion in the image \irt{} with respect to the reference image \ilt{} is modeled at each pyramid level $l$ as an occlusion map ${_l}\vectr{o}_R^{t}(\vectr{x})$ where ${_l}\vectr{o}_R^{t}: \Omega \to [0, 1]$ and $\Omega$ denotes the image plane. Here 0 corresponds to occluded pixels while 1 corresponds to visible pixels. Since each pixel value is continuous, this is a soft occlusion map from which a hard occlusion map can be obtained by thresholding it appropriately to have discrete 0 and 1 values.

This soft occlusion map is incorporated into \network{} by multiplying it pixel-wise (by broadcasting along the channel dimension) with the corresponding warped features to result in masked features ${_l}\vectr{f}_{R}^{t}$:
\begin{equation}
    {_l}\vectr{f}_{R}^{t}(\vectr{x}) = {_l}\vectr{c}_{R}^{t}(\vectr{x}) \cdot {_l}\vectr{o}_{R}^{t}(\vectr{x})
\end{equation}
This has the effect of masking out occluded pixels from ${_l}\vectr{w}_{R}^{t}$, leaving only the non-occluded areas. These masked warped features are then used to construct the cost volume. The occluded pixels having been masked to 0 results in the cost for these pixels to also be computed as 0. In the cost volume operation, a higher cost means a higher degree of matching between two pixels. Thus, a cost of 0 as computed for the occluded areas reflects on no matching at all, which is semantically correct, since a pixel occluded in one image must not match with any other pixel in another image.

Due to the coarse-to-fine estimation approach adopted by \network, the occlusion model is also a multi-scale mechanism. A separate occlusion map is predicted at each pyramid level $l$ for each of the warped image features (${_l}\vectr{w}_{R}^t$, ${_l}\vectr{w}_{L}^{t+1}$, ${_l}\vectr{w}_{R}^{t+1}$), which masks out occluded areas for feeding to the respective cost volume at that level.

\textbf{Learning Occlusions.} For predicting occlusion, we train a separate network $\mathcal{O}^l$ at each scale $l$ which maps the depthwise stacked feature maps \clt{} and ${_l}\vectr{w}_{R}^{t}$ to the soft occlusion map ${_l}\vectr{o}_{R}^{t}$. These stacked feature maps are used as input to the network because they provide sufficient information to predict occlusion: Regions without occlusion would have similar features in the feature map \clt{} and the warped features ${_l}\vectr{w}_{R}^{t}$; whereas pixels which are visible in \clt{} but occluded in \crt{} would have a dissimilarity in the features in ${_l}\vectr{w}_{R}^{t}$, which can enable the network to predict such pixels as occluded.

The design of the $\mathcal{O}^l$ network consists of six layers of convolution, all of which employ a kernel of size $3\times3$, a stride of 1 and a padding of 1. The channel dimensions of the occlusion estimator network in each layer are 128, 96, 64, 32, 16 and 1 respectively. The last layer uses sigmoid as an activation function to ensure that the occlusion predictions are in the range $[0, 1]$. All other layers use the leaky ReLU activation function.

This network is inserted in the \network{} pipeline after the warping operation and before the cost volume construction stage at each pyramid level. After warping the three images towards the reference image, the reference image features are stacked with each of the other feature maps as [\clt, ${_l}\vectr{w}_{R}^t$], [\clt, ${_l}\vectr{w}_{L}^{t+1}$], [\clt, ${_l}\vectr{w}_{R}^{t+1}$] and sequentially fed as input to the occlusion estimator network $\mathcal{O}^l$ at that pyramid level to obtain the occlusion maps ${_l}\vectr{o}_{R}^t$, ${_l}\vectr{o}_{L}^{t+1}$, ${_l}\vectr{o}_{R}^{t+1}$ respectively. Since the estimation of occlusion for each of the 3 images requires learning the same underlying task (that of learning to match features in the stacked image feature maps) a single occlusion estimator network $\mathcal{O}^l$ is used at each pyramid level to predict occlusion for each of the three images separately.

To maintain consistency of occlusion predictions across scales, every occlusion network $\mathcal{O}^l$ (except the network at the highest pyramid level) receives as additional input from the network $\mathcal{O}^{l+1}$ (of the upper pyramid level), the output features ${_{l+1}}\vectr{g}$ of its penultimate convolutional layer and the corresponding predicted occlusion map $\vectr{o}$. That is, $({_{l+1}}\vectr{g}_R^{t},\ {_{l+1}}\vectr{o}_R^{t})$, $({_{l+1}}\vectr{g}_L^{t+1},\ {_{l+1}}\vectr{o}_L^{t+1})$ and $({_{l+1}}\vectr{g}_R^{t+1},\ {_{l+1}}\vectr{o}_R^{t+1})$ are the additional inputs to predict ${_{l}}\vectr{o}_R^{t}$, ${_{l}}\vectr{o}_L^{t+1}$ and ${_{l}}\vectr{o}_R^{t+1}$ respectively.

The occlusion estimator networks at each level can learn to predict the occlusion weights based on the degree of similarity between the reference image features and the warped features. These weights are used to mask the incorrect matching costs in the cost volume, which results in more robust scene flow estimations. Thus, the network \textit{supervises itself} while learning to estimate occlusions, with the goal of improving scene flow estimates (minimizing the error on scene flow predictions) without any labeled occlusion data. Note that the training of \network{} requires ground truth scene flow data; only the estimation of the occlusion maps is self-supervised.

\textbf{Cost Volume.} We compute a cost volume using the reference image features \clt{} and the masked warped features ${_l}\vectr{f}_{R}^t$, ${_l}\vectr{f}_{L}^{t+1}$ and ${_l}\vectr{f}_{R}^{t+1}$. In contrast, PWC-Net computed the cost volume using simply the warped features ${_l}\vectr{w}_{R}^t$, ${_l}\vectr{w}_{L}^{t+1}$ and ${_l}\vectr{w}_{R}^{t+1}$, which made its predictions susceptible to problems caused by occlusions.

We construct only a partial cost volume by limiting the search range to a maximum displacement of $d_{\text{max}}$ pixels around each pixel. For the 1D cost volume operation (between (\clt, ${_l}\vectr{f}_{R}^t$)), we search for matches in the horizontal dimension only along the epipolar line, while for the 2D cost volume (between (\clt, ${_l}\vectr{f}_{L}^{t+1}$) and (\clt, ${_l}\vectr{f}_{R}^{t+1}$)) we search in 2D space. We then organize the resulting cost volume as a 3D array of dimensions $H \times W \times D$ and $H \times W \times D^2$ for the 1D and 2D cost volumes respectively where $H$ and $W$ are the height and width of the feature maps respectively and $D = 2d_{\text{max}}+1$.

The matching cost in the cost volume is computed as the correlation between the feature vectors of two pixels. Consider \clt, ${_l}\vectr{f}_{R}^t: \rm \Omega \mapsto I\!R^c$, where $\Omega$ is the set of image pixel vectors and $c$ is the number of channels of the feature maps. Then the correlation between two patches centred at pixels $\vectr{x}_1$ and $\vectr{x}_2$ is computed as a vector of dimensionality $D^2$ where each individual pixel cost is given by:
\begin{equation}
    {_l}\vectr{C}_R^t(\vectr{x}_1, \vectr{x}_2 | \vectr{q}) = \frac{({_l}\vectr{c}_{L}^t(\vectr{x}_1))^\top {_l}\vectr{f}_{R}^t(\vectr{x}_2 + \vectr{q})}{c},
\end{equation}
where $\vectr{q} \in \{(q_0, q_1)^\top: q_0, q_1 \in [-d_{\text{max}}, d_{\text{max}}]\}$.

\textbf{Scene Flow Prediction.} At every pyramid level $l$, we train a CNN $\mathcal{S}^l$ to estimate scene flow using the masked cost volume described above. The primary input to this network consists of the three warped and masked cost volumes corresponding to ${_l}\vectr{f}_{R}^t$, ${_l}\vectr{f}_{L}^{t+1}$ and ${_l}\vectr{f}_{R}^{t+1}$ stacked one over the other along the channel dimension. The architecture of this network is similar to the occlusion estimator network: It consists of six layers of convolution filters with a kernel size of $3\times3$, a stride of 1 and a padding of 1. The channel dimensions of each layer are 128, 128, 96, 64, 32 and 4 respectively. Each layer employs the leaky ReLU activation, except the last layer which does not use any activation function to facilitate it to predict continuous scene flow values.

Similar to the occlusion estimator network, we pass as additional input to $\mathcal{S}^l$ (at each pyramid level $l$ except the top), the features $\vectr{h}^{l+1}$ from the penultimate convolutional layer of the network $\mathcal{S}^{l+1}$ (of the upper pyramid level) as well as its corresponding scene flow prediction $\vectr{s}^{l+1}$. This maintains consistency across the pyramid levels and allows the entire framework to perform multi-scale reasoning.

\begin{figure*}
	\centering
	\begin{subfigure}[b]{0.24\linewidth}
		\includegraphics[width=\linewidth]{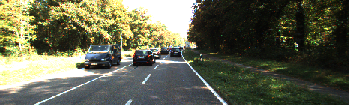}
		\caption{\ilt}
		\label{fig:im0}
	\end{subfigure}
	\begin{subfigure}[b]{0.24\linewidth}
		\includegraphics[width=\linewidth]{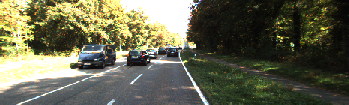}
		\caption{\irt}
		\label{fig:im1}
	\end{subfigure}
	\begin{subfigure}[b]{0.24\linewidth}
		\includegraphics[width=\linewidth]{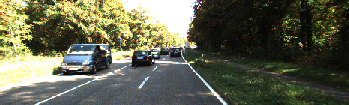}
		\caption{\ilT}
		\label{fig:im2}
	\end{subfigure}
	\begin{subfigure}[b]{0.24\linewidth}
		\includegraphics[width=\linewidth]{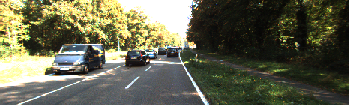}
		\caption{\irT}
		\label{fig:im3}
	\end{subfigure}
	\begin{subfigure}[b]{0.3\linewidth}
		\includegraphics[width=\linewidth]{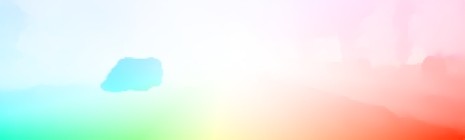}
		\caption{Optical flow prediction $\vectr{u}_\Theta$.}
		\label{fig:occ-flow}
	\end{subfigure}
    \begin{subfigure}[b]{0.3\linewidth}
		\includegraphics[width=\linewidth]{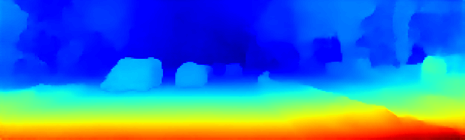}
		\caption{First disparity prediction $d_{0\Theta}$.}
		\label{fig:d0}
	\end{subfigure}
    \begin{subfigure}[b]{0.3\linewidth}
		\includegraphics[width=\linewidth]{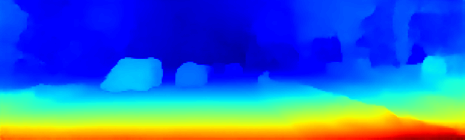}
		\caption{Second disparity prediction $d_{1\Theta}$.}
		\label{fig:d1}
	\end{subfigure}
    \begin{subfigure}[b]{0.3\linewidth}
		\includegraphics[width=\linewidth]{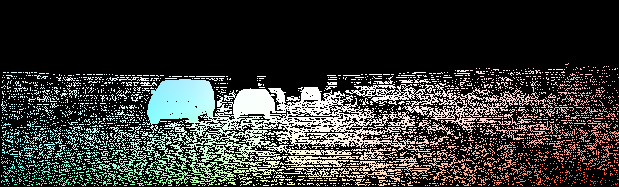}
		\caption{Optical flow ground truth $\vectr{u}_\mathrm{GT}$.}
		\label{fig:gt-flow}
	\end{subfigure}
    \begin{subfigure}[b]{0.3\linewidth}
		\includegraphics[width=\linewidth]{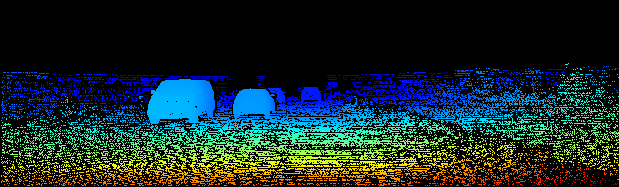}
		\caption{First disparity ground truth $d_{0\mathrm{GT}}$.}
		\label{fig:gt-d0}
	\end{subfigure}
    \begin{subfigure}[b]{0.3\linewidth}
		\includegraphics[width=\linewidth]{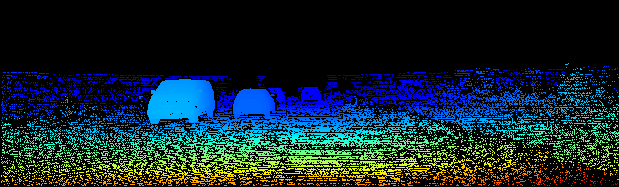}
		\caption{Second disparity ground truth $d_{1\mathrm{GT}}$.}
		\label{fig:gt-d1}
	\end{subfigure}
    \begin{subfigure}{0.3\linewidth}
        \includegraphics[width=\linewidth]{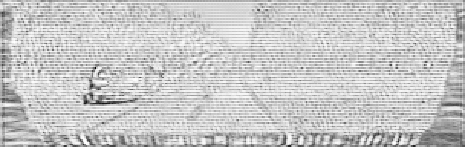}
        \caption{Occlusion mask for optical flow $\vectr{o}_L^{t+1}$.}
        \label{fig:flow-occ}
    \end{subfigure}
    \begin{subfigure}{0.3\linewidth}
        \includegraphics[width=\linewidth]{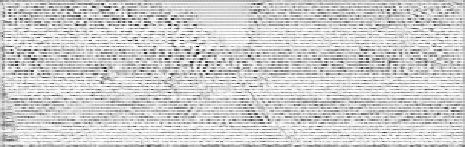}
        \caption{Occlusion mask for first disparity $\vectr{o}_R^t$.}
        \label{fig:d0-occ}
    \end{subfigure}
    \begin{subfigure}{0.3\linewidth}
        \includegraphics[width=\linewidth]{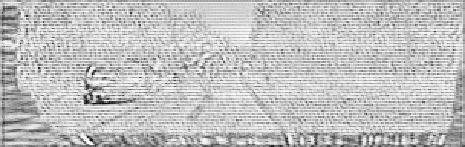}
        \caption{Occlusion mask for second disparity $\vectr{o}_R^{t+1}$.}
        \label{fig:d1-occ}
    \end{subfigure}
	\begin{subfigure}{0.3\linewidth}
		\includegraphics[width=\linewidth]{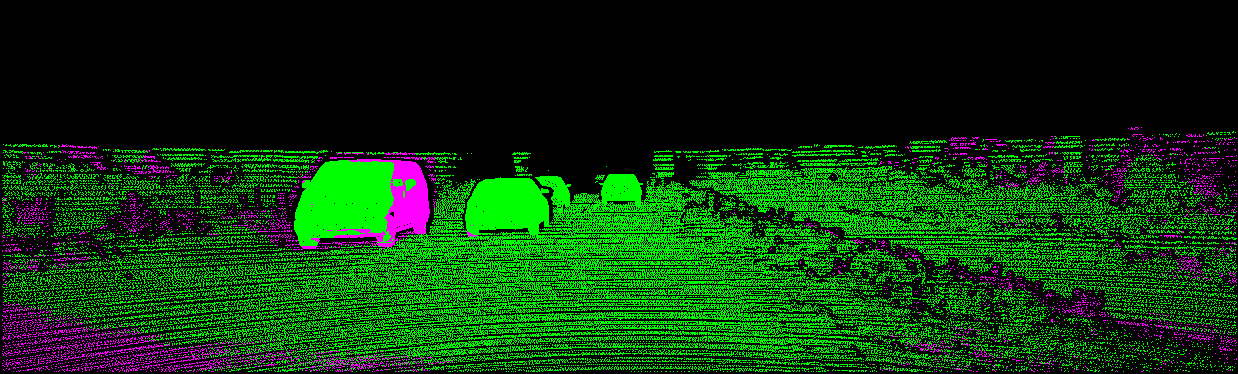}
		\caption{Binary error map for optical flow.}
		\label{fig:flow-error}
	\end{subfigure}
	\begin{subfigure}{0.3\linewidth}
		\includegraphics[width=\linewidth]{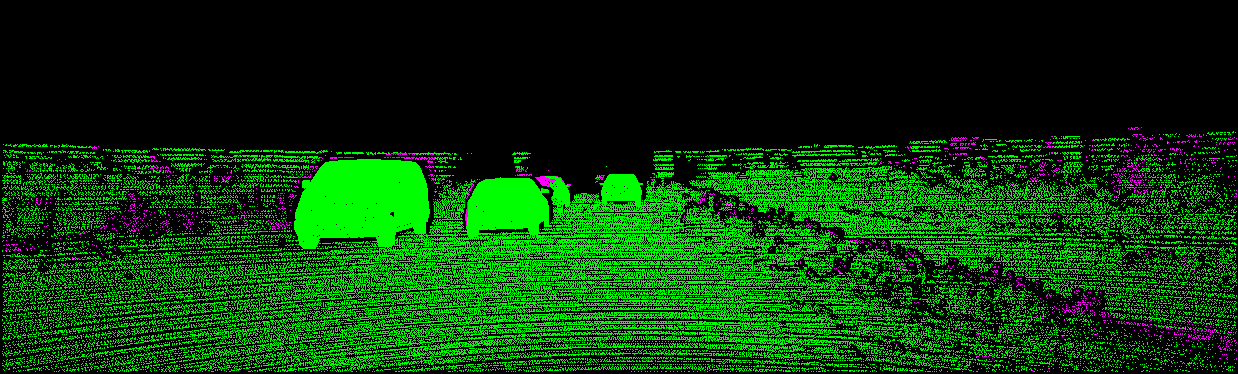}
		\caption{Binary error map for first disparity.}
		\label{fig:d0-error}
	\end{subfigure}
	\begin{subfigure}{0.3\linewidth}
		\includegraphics[width=\linewidth]{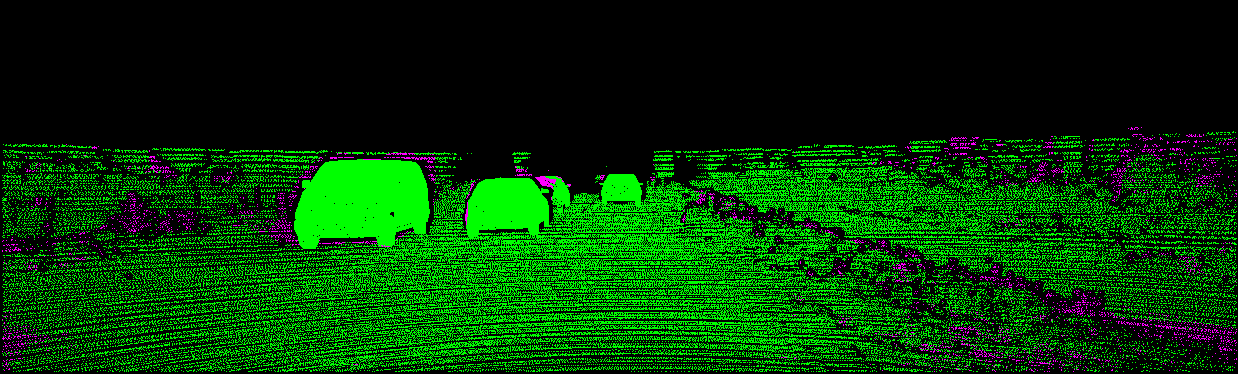}
		\caption{Binary error map for second disparity.}
		\label{fig:d1-error}
	\end{subfigure}
	\caption{Visualization of predictions of \network{} with occlusion masks and ground truth on a validation sample from KITTI. In the error maps \subref{fig:flow-error}, \subref{fig:d0-error} \subref{fig:d1-error}, pixels which contribute to the KITTI outlier error are coloured in magenta, while those that do not are coloured green.}
	\label{fig:visualization}
\end{figure*}

\textbf{Context Network.} Similar to PWC-Net, our \network{} employs a CNN with dilated convolutional layers to refine the flow estimation. The input to this network comprises the flow estimate $\vectr{s}^2$ and the last feature map $\vectr{f}^2$ from the scene flow estimator $\mathcal{S}^2$ of the lowest pyramid level. This context network consists of 7 convolutional layers with $3\times3$ filters with a stride and padding of 1. The number of filters at each layer are 128, 128, 128, 96, 64, 32 and 4 respectively. The dilation parameters used at each layer are 1, 2, 4, 8, 16, 1 and 1 respectively. All layers use the leaky ReLU activation, except the last which does not employ any activation.

This network outputs a residual flow $\delta\vectr{s}^2$ which is added to the scene flow prediction $\vectr{s}^2$ to obtain the final prediction $\vectr{\^{s}}(\vectr{x}) = \vectr{s}^2(\vectr{x}) + \delta\vectr{s}^2(\vectr{x})$.

\textbf{Loss Function.} To start the coarse-to-fine estimation scheme of \network, we initialize the prediction at a hypothetical level 7 of the pyramid $\vectr{s}^7$ with zeros. This has the effect that the features at the topmost level ($l=6$) of the pyramid are not warped at all. Thus, the cost volume and occlusions are computed directly using the original feature maps. The rest of the pipeline progresses as described in the previous sections.

We employ a multi-scale weighted loss function with intermediate supervision which penalizes losses at each level of the pyramid. Let $\Theta$ denote the set of all trainable parameters in the entire network and let $\vectr{s}_\mathrm{GT}^l$ be the ground truth scene flow field subsampled to the resolution of pyramid level $l$, leading to the loss function
\begin{equation}
    \begin{split}
        \mathcal{L}(\Theta) = &\underbrace{\sum_{l=3}^6 \alpha_l \sum_{\vectr{x}} \lvert \vectr{s}_\Theta^l(\vectr{x}) - \vectr{s}_\mathrm{GT}^l(\vectr{x}) \rvert_2}_\text{pyramid levels except the lowest}\\
        + &\underbrace{\alpha_{2} \sum_{\vectr{x}} \lvert \vectr{\^{s}}_\Theta(\vectr{x}) - \vectr{s}_\mathrm{GT}^{2}(\vectr{x}) \rvert_2}_\text{bottom pyramid level $2$} + \underbrace{\gamma \lvert \Theta \rvert_2}_\text{\tiny{L2 regularization}},
    \end{split}
\end{equation}
where $\lvert \text{} \cdot \text{} \rvert_2$ denotes the L2 norm of a vector.

This enables the entire \network{} model: The feature pyramid network, the occlusion mechanism with its occlusion estimator networks at different pyramid levels, the scene flow estimator networks at each pyramid level, and the context network to be trained in an end-to-end manner.

\begin{table*}
    \centering
    \caption{Experimental results of \network{}. We show endpoint error (EPE [px]) and KITTI outlier error (KOE [\%]) on training, validation, and test set for KITTI and FlyingThings3D. We evaluate different components of our contribution and compare to end-to-end scene flow networks from previous work.}
    \label{table:eval}
    \begin{adjustbox}{center}
    %\resizebox{1\linewidth}{!}{%
	    \begin{tabular}{l c c c c c c c c c c c c c c}
	        \toprule
	        & \hphantom{-} & \multicolumn{6}{c}{\textbf{FlyingThings3D}} & \hphantom{-} & \multicolumn{6}{c}{\textbf{KITTI}}\\
	        &  & \multicolumn{2}{c}{\textbf{Training}} & \multicolumn{2}{c}{\textbf{Validation}} & \multicolumn{2}{c}{\textbf{Testing}} &  & \multicolumn{2}{c}{\textbf{Training}} & \multicolumn{2}{c}{\textbf{Validation}} & \multicolumn{2}{c}{\textbf{Testing}}\\
	        \textbf{Architecture} &  & EPE & KOE & EPE & KOE & EPE & KOE &  & EPE & KOE & EPE & KOE & EPE & KOE\\
	        \midrule
	        \network{} (\textit{basic}) &  & 8.25 & 25.15 & 9.79 & 25.01 & 23.38 & 26.01 &  & 1.97 & 6.09 & 3.71 & 13.6 & -- & --\\
	        \network{} + FPN &  & 6.17 & 19.89 & 8.40 & 20.35 & \textbf{21.86} & 21.22 &  & \textbf{1.76} & \textbf{5.32} & 3.39 & 13.97 & -- & --\\
	        \network{} + FPN + Occ &  & \textbf{5.86} & \textbf{18.30} & \textbf{8.06} & \textbf{18.93} & 22.01 & \textbf{19.90} &  & 1.85 & 5.69 & \textbf{3.22} & \textbf{12.55} & -- & \textbf{15.69}\\
	        \midrule
	        SceneFlowNet \cite{mayer2016large} & & -- & -- & 11.24 & -- & -- & -- &  & -- & -- & -- & -- & -- & --\\
	        Occ-SceneFlow \cite{occlusionssceneflow} &  & -- & -- & -- & -- & -- & -- &  & -- & -- & -- & -- & -- & 11.34\\
	        \bottomrule
	    \end{tabular}
	\end{adjustbox}
    %}
\end{table*}

\begin{table*}[t]
	\centering
	\caption{A snapshot of the KITTI scene flow benchmark's leaderboard at the time of submission. Our method is the most efficient in terms of runtime.}
	\label{tab:ranking}
	\begin{tabular}{c c c c c c c c c c c c c c}
		\toprule
		{\bf Method} & {\bf D1-bg} & {\bf D1-fg} & {\bf D1-all} & {\bf D2-bg} & {\bf D2-fg} & {\bf D2-all} & {\bf Fl-bg} & {\bf Fl-fg} & {\bf Fl-all} & {\bf SF-bg} & {\bf SF-fg} & {\bf SF-all} & {\bf Runtime}\\
		\midrule
		ISF \cite{behl2017bounding} & 4.12 & \textbf{6.17} & 4.46 & \textbf{4.88} & \textbf{11.34} & \textbf{5.95} & 5.40 & \textbf{10.29} & \textbf{6.22} & \textbf{6.58} & \textbf{15.63} & \textbf{8.08} & 600 s \\
		PRSM \cite{vogel2015PRSM} & \textbf{3.02} & 10.52 & \textbf{4.27} & 5.13 & 15.11 & 6.79 & \textbf{5.33} & 13.40 & 6.68 & 6.61 & 20.79 & 8.97 & 300 s \\
		OSF+TC \cite{neoral2017object} & 4.11 & 9.64 & 5.03 & 5.18 & 15.12 & 6.84 & 5.76 & 13.31 & 7.02 & 7.08 & 20.03 & 9.23 & 3000 s \\
		OSF 2018 \cite{menze2018osf} & 4.11 & 11.12 & 5.28 & 5.01 & 17.28 & 7.06 & 5.38 & 17.61 & 7.41 & 6.68 & 24.59 & 9.66 & 390 s \\
		SSF \cite{ren2017cascaded} & 3.55 & 8.75 & 4.42 & 4.94 & 17.48 & 7.02 & 5.63 & 14.71 & 7.14 & 7.18 & 24.58 & 10.07 & 300 s \\
		OSF \cite{menze2015object} & 4.54 & 12.03 & 5.79 & 5.45 & 19.41 & 7.77 & 5.62 & 18.92 & 7.83 & 7.01 & 26.34 & 10.23 & 3000 s\\
		FSF+MS \cite{taniai2017fast} & 5.72 & 11.84 & 6.74 & 7.57 & 21.28 & 9.85 & 8.48 & 25.43 & 11.30 & 11.17 & 33.91 & 14.96 & 2.7 s \\
		\textbf{PWOC-3D (ours)} & 4.19 & 9.82 & 5.13 & 7.21 & 14.73 & 8.46 & 12.40 & 15.78 & 12.96 & 14.30 & 22.66 & 15.69 & \textbf{0.13 s} \\
		CSF \cite{lv2016continuous} & 4.57 & 13.04 & 5.98 & 7.92 & 20.76 & 10.06 & 10.40 & 25.78 & 12.96 & 12.21 & 33.21 & 15.71 & 80 s \\
		SFF \cite{schuster2018sceneflowfields} & 5.12 & 13.83 & 6.57 & 8.47 & 21.83 & 10.69 & 10.58 & 24.41 & 12.88 & 12.48 & 32.28 & 15.78 & 65 s \\
		PR-Sceneflow \cite{vogel2013PRSF} & 4.74 & 13.74 & 6.24 & 11.14 & 20.47 & 12.69 & 11.73 & 24.33 & 13.83 & 13.49 & 31.22 & 16.44 & 150 s\\
		\bottomrule
	\end{tabular}
\end{table*}

\section{EXPERIMENTS AND RESULTS}

\textbf{Datasets.} The primary focus of this work is to estimate scene flow for automotive applications, thus making the KITTI dataset \cite{menze2015object} a natural choice. However, KITTI provides only 200 training sequences (with sparse ground truth only), which is not sufficient to train a deep neural network. To overcome this problem, we first pretrain \network{} on the synthetic (but large) FlyingThings3D dataset \cite{mayer2016large}, and then finetune this model on KITTI. This transfer learning approach helps avoiding the network from being overfit on KITTI. We train \network{} for 760 epochs on FlyingThings3D and for 125 epochs on KITTI.

\textbf{Training Details.} We use the photometric data augmentation strategy of FlowNet \cite{flownet}, combined with random vertical flipping (the latter only for FlyingThings3D, and not for KITTI). Since FlyingThings3D also provides bidirectional scene flow annotations, we utilise this by random temporal flipping of the training sample from (\ilt, \irt, \ilT, \irT) to (\ilT, \irT, \ilt, \irt). We refrain from geometric transformations such as rotation, translation, etc which would destroy the epipolar constraint for disparity estimation across the stereo pairs.

The hyperparameter $d_{\text{max}}$ in the cost volume layer is set to 4. The weights used in the loss function $\alpha_2, \alpha_3, \dots, \alpha_6$ are 0.32, 0.08, 0.02, 0.01 and 0.005 respectively. The regularization parameter $\gamma$ is set to 0.
Further, the ground truth scene flow is downscaled by 20 as in \cite{flownet}, and is downsampled to different resolutions to compute the training signal at different scales. During inference, all scene flow predictions are made at the input resolution.
We use the Adam \cite{kingma2015adam} optimizer to train \network{} with the default setting of hyperparameters as recommended in \cite{kingma2015adam}. We use a learning rate of $\lambda = 10^{-4}$.

\textbf{Quantitative Analysis.} 
As evident from \cref{table:eval}, the \network{} model with the occlusion mechanism and the improved feature pyramid is the best performing network among all the variants. This is due to its well-localized and semantically strong features and its ability to mask the incorrect matching costs from the cost volume, thus preventing them from having adverse effects on the reasoning of the network.
\Cref{table:eval} also shows the comparison to the only two other end-to-end scene flow networks. \network{} outperforms \cite{mayer2016large} on FlyingThings3D in terms of endpoint error and is 48 times smaller than the network from \cite{occlusionssceneflow} (\network{} has only 8,046,625 trainable weights).

As depicted in \cref{fig:visualization}, the occlusion maps contain clear signs of the masking effect. Specifically, the occlusion map for $d_0$ shown in \cref{fig:d0-occ} contains areas occluded only along the horizontal epipolar line, while the maps for optical flow and $d_1$ in \cref{fig:flow-occ,fig:d1-occ} respectively model the occlusion caused due to the ego-motion of the camera in addition to the occlusion arising because of motion. This demonstrates the significant impact of our occlusion mechanism in autonomous driving scenarios.

The network with the FPN shows improvement in performance over the network without. The endpoint error of all the networks on the test split of FlyingThings3D \cite{mayer2016large} is higher than that on the validation split as visible in \cref{table:eval}. This is because in the test set, FlyingThings3D contains a set of objects and backgrounds which are disjoint from the training set. Thus, the samples are considerably different from those that the networks have been trained on.

Another interesting result is that among all variants in \cref{table:eval}, the difference between the validation and training errors is the lowest for \network{} with the feature pyramid connections and the occlusion mechanism, particularly on KITTI where the training data is very limited. Thus, our occlusion reasoning scheme also helps reduce overfitting.

\Cref{tab:ranking} shows the ranking of our method among the top 10 published methods on the KITTI scene flow benchmark's leaderboard at the time of our submission. As visible, \network{} has a significantly lower runtime (0.13 s per frame on a GeForce GTX 1080 Ti) than all other methods, thus making it suitable for real-time applications. Among the listed methods, \network{} is the only approach with that property. Further, our approach is especially accurate in the important foreground regions of moving objects.
In summary, \network{} has a unique mixture of characteristics with competitive accuracy, small network size, and low runtime.

\section{CONCLUSION}
In this paper we proposed \network, a novel end-to-end CNN pipeline to predict scene flow (optical flow and stereo disparity jointly) directly from stereo image sequences. Our approach was significantly more efficient than earlier classical approaches, and much more accurate than variational methods. Moreover, unlike most previous techniques, \network{} does not make any assumptions about the consistency or smoothness of motion, or the rigidity of objects. This makes our method more general and applicable to realistic scenarios where such assumptions do not hold, e.g.\ highly dynamic road scenes.

Moreover, \network{} employs special constructs such as pyramid processing, warping and occlusion reasoning to tackle common challenges in scene flow like large motion and occlusions. In this regard, we proposed a novel self-supervised scheme to estimate occlusion from images without any labeled occlusion data. \network{} demonstrates competitive results on the KITTI benchmark and the FlyingThings3D dataset. Notably, our method has significantly fewer parameters than contemporary methods and achieves second place on KITTI among end-to-end deep learning methods with 48 times fewer parameters than the top-performing method.
\network{}, along with our self-supervised occlusion scheme, can be combined with an unsupervised reconstruction loss (similar to \cite{meister2018unflow,wang2018occlusion}) to result in a fully self-supervised end-to-end CNN which predicts scene flow. This requires a detailed and comprehensive study and is left for future work.

\section*{Acknowledgment}
\vspace*{-5pt}
This work was partially funded by the Federal Ministry of Education and Research Germany as part of the project VIDETE (01IW18002).
\vspace*{-1pt}

\addtolength{\textheight}{-12cm}   % This command serves to balance the column lengths

%%%%%%%%%%%%%%%%%%%%%%%%%%%%%%%%%%%%%%%%%%%%%%%%%%%%%%%%%%%%%%%%%%%%%%%%%%%%%%%%

\bibliographystyle{IEEEtran}
\bibliography{literature}

\end{document}